\pdfoutput=1

\documentclass[11pt]{article}

\usepackage[preprint]{acl}

\usepackage{times}
\usepackage{latexsym}

\usepackage[T1]{fontenc}

\usepackage[utf8]{inputenc}

\usepackage{microtype}

\usepackage{inconsolata}

\usepackage{graphicx}

\usepackage{amsmath}
\usepackage{amsfonts}
\usepackage{booktabs}
\usepackage{makecell}
\usepackage{multirow}
\usepackage{subfigure}
\usepackage{float}
\usepackage{circledsteps}
\usepackage{colortbl}
\usepackage{soul}

\title{HERA: Improving Long Document Summarization using Large Language Models with Context Packaging and Reordering}

\author{
 \textbf{Taiji Li\textsuperscript{1}},
 \textbf{Hao Chen\textsuperscript{2}},
 \textbf{Fei Yu\textsuperscript{2}},
 \textbf{Yin Zhang\textsuperscript{1}}\thanks{Corresponding Author: Yin Zhang.},
\\
 \textsuperscript{1} College of Computer Science and Technology, Zhejiang University
\\
 \textsuperscript{2} Ant Group, China
\\
 \small{
   \texttt{\{litaiji, zhangyin98\}@zju.edu.cn, chuhu.ch@antgroup.com, feiyu.fyyu@gmail.com }
 }
}

\begin{document}
\maketitle
\begin{abstract}
Despite the rapid growth of context length of large language models (LLMs) , LLMs still perform poorly in long document summarization. An important reason for this is that relevant information about an event is scattered throughout long documents, and the messy narrative order impairs the accurate understanding and utilization of LLMs for long documents. To address these issues, we propose a novel summary generation framework, called \textbf{HERA}. Specifically, we first segment a long document by its semantic structure and retrieve text segments about the same event, and finally reorder them to form the input context. We evaluate our approach on two long document summarization datasets. The experimental results show that HERA outperforms foundation models in ROUGE, BERTScore and faithfulness metrics, while HERA does not require additional fine-tuning and resources.
\end{abstract}

\section{Introduction}

Long document summarization aims to generate fluent, concise and faithful summaries of long texts such as government documents, scientific papers and books. Benefiting from the development of LLMs, long document summarization is no longer a difficult task. Many newly released LLMs, such as Claude 3 and Gemini 1.5, are already able to process millions of tokens at a time. However, the performance of LLMs significantly decreases as the length of context grows \citep{dong-etal-2024-bamboo-comprehensive, hsieh2024ruler}. More seriously, the summaries generated by LLMs often contain a lot of content that is irrelevant to or contradicts the original document \citep{li-etal-2023-halueval, tam-etal-2023-evaluating}, which undermines the reliability and usability of the text summarization.

\begin{table}[t]
    \centering
    \resizebox{\linewidth}{!}{
    \begin{tabular}{|p{\linewidth}|}
    \hline
        \textbf{Article}: 

        \Circled{1} The 2024 UEFA Champions League final was held at \textcolor{red}{Wembley Stadium} in London, England, \textcolor{red}{on 1 June 2024}, between German club Borussia \textcolor{red}{Dortmund} and Spanish club \textcolor{red}{Real Madrid}. ...
        
        \Circled{6} For Borussia Dortmund, this was their third UEFA Champions League final appearance, and Real Madrid played in a record-extending 18th European Cup/UEFA Champions League final, and their second in three years. They previously \textcolor{red}{won 14 finals} and lost three. ...
        
        \Circled{8} Real Madrid defeated Dortmund \textcolor{red}{2-0} to become the team with \textcolor{red}{the most European Cup/UEFA Champions League titles of all time}. ...\\
    \hline
        \textbf{Summaries:}
        
        On June 1, 2024, Real Madrid defeated Dortmund 2-0 at Wembley Stadium to win a record-extending 15th title. \\
    \hline
    \end{tabular}}
    \caption{The information in the summary is spread out in paragraphs far apart.}
    \label{tab:diagram}
\end{table}

\begin{figure*}[t]
  \includegraphics[width=\linewidth]{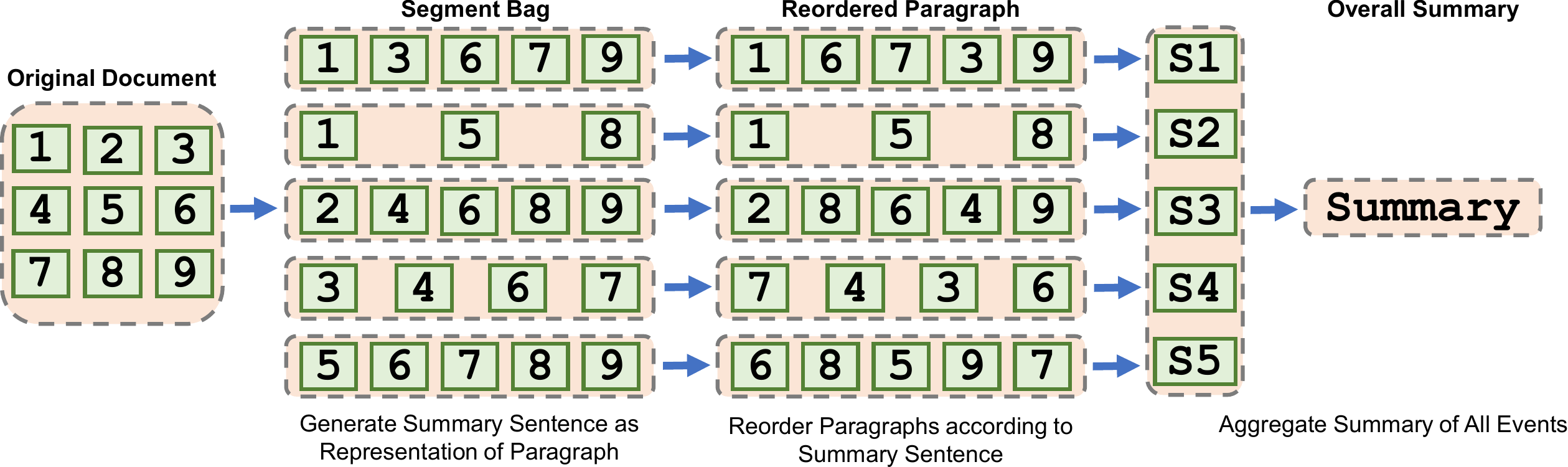}
  \caption {The pipeline of our proposed \textbf{HERA} approach.}
  \label{fig:architecture}
\end{figure*}

Many studies explore the mechanism why the performance of LLMs degrades in long-context scenarios. \citet{10.1162/tacl_a_00638} observe that LLMs prefer to extract information at the beginning or end of the context and ignore the content in the middle. The Needle-in-a-Haystack \citep{zhao2024longagent} shows that LLMs have difficulty finding the required information in massive texts. \citet{wu2024easily} and \citet{du2024context} demonstrates that LLMs can be easily distracted by these irrelevant yet misleading contents. In addition, many works \citep{kumar-talukdar-2021-reordering, lu-etal-2022-fantastically, wu-etal-2023-self, zhang-etal-2023-reading} have revealed that the reading order of LLMs has a significant impact on their understanding and utilization of context. Therefore, as shown in Table \ref{tab:diagram}, the key to improving the performance of LLMs in long document summarization is \emph{how to extract useful information and arrange it in the correct narrative order}.

To address these problems, we propose \textbf{HERA}, a long document summary generation framework via context packaging and reordering. The document is divided into small text segments by paragraph, then we retrieve the segments about the one event and we use LLM to arrange in semantic order to form the input. After that, LLM uses the packed context to generate a summary of the event. Finally, the summaries of several important events are combined as the summary of the entire document. The packaged context only retains the content related to the event, preventing LLMs from being misled by irrelevant information, while the reordered context promotes LLMs to understand the information more accurately. 

We evaluate the effectiveness of HERA applied to four advanced LLMs, LLaMA 2, LLaMA 3, Gemini 1.5 and GPT-4, on two widely-used long document summarization datasets arXiv and PubMed \citep{cohan-etal-2018-discourse}. Extensive experiments show that our proposed method significantly improves the performance of diverse foundation models on long document summarization, and achieves state-of-the-art performance on both fluency and faithfulness metrics.  Besides, we conduct ablation studies to further investigate
why HERA works. Our contributions are the followings: (1) We propose a novel summary generation framework, \textbf{HERA}, that improves long document summarization via context packaging and reordering without requiring additional training and resources. (2) We demonstrate that HERA improves the fluency and faithfulness of long document summarization and can be applied to diverse LLMs. (3) We conduct ablation experiments to investigate the effectiveness of context packaging and reordering, and their impact on performance.

\section{Approach}

\paragraph{Context Packaging} HERA divides the original document into a series of text segments by paragraphs, and use a small summarization model to generate  sentence-long local summaries for these text segments as their keys for retrieval and reordering. Because LLMs have better information retrieval and ranking capabilities than supervised methods \citep{sun-etal-2023-chatgpt}, HERA uses an LLM to retrieve paragraphs related to each event in turn. If these partial abstracts are still too long, HERA will split them into several parts for easier retrieval. then selects the top-ranked segments as the relevant paragraphs for that event, combining them into a separate segment bag.

\paragraph{Context Reordering} There are many models \citep{lai-etal-2021-improving, Zhu_Zhou_Nie_Liu_Dou_2021, ghosal-etal-2021-stack, jia-etal-2023-sentence} for sentence ordering, which make the final output logically smooth and clear. Most of models use graph neural networks or Transformers to extract co-occurrence relationships and semantic connections between sentences. HERA uses the current state-of-the-art sentence reordering model NAON \citep{bin-etal-2023-non-autoregressive} to order the paragraphs in the segment bag, and HERA use the summary sentence of each paragraph as their representative to speed up the sorting. Finally, HERA sort the corresponding paragraphs according to the order of the summary sentences. HERA uses LLM to generate summaries for the reordered segment bag and again uses LLM to aggregate these summaries into an overall summary of the original document. Figure \ref{fig:architecture} illustrates the process of HERA for generating summaries for long documents.

\section{Experiments}

\begin{table*}[t]
\centering
\resizebox{\linewidth}{!}{
    \begin{tabular}{l|cccc|cc|cccc|cc}
    \toprule
         & \large{\textbf{R-1}} & \large{\textbf{R-2}} & \large{\textbf{R-L}} & \large{\textbf{BS}} & \large{\textbf{FC}} & \large{\textbf{SC}} & \large{\textbf{R-1}} & \large{\textbf{R-2}} & \large{\textbf{R-L}} & \large{\textbf{BS}} & \large{\textbf{FC}} & \large{\textbf{SC}} \\ 
    \midrule
        \large{\textbf{Model}} & \multicolumn{6}{c|}{\large{\textbf{arXiv}}} & \multicolumn{6}{c}{\large{\textbf{PubMed}}}\\
    \midrule
        FactorSum & 48.34 & 20.57 & 42.82 & 88.42 & 63.85 & 59.24 & 47.72 & 20.61 & 43.95 & 83.56 & 69.54 & 63.51\\
        Lodoss & 48.52 & 20.79 & 42.91 & 88.73 & 72.45 & 68.34 & 49.42 & 23.86 & 44.82 & 88.75 & 78.46 & 75.26\\
        \hline
        LLaMA 2 & 39.26 & 14.31 & 34.63 & 79.48 & 56.28 & 53.17 & 41.63 & 17.52 & 37.18 & 77.53 & 62.74 & 59.32\\ \rowcolor{orange!30}
        \textbf{LLaMA 2 + HERA} & 46.75 & 18.83 & 40.67 & 84.51 & 69.18 & 68.32 & 47.25 & 21.59 & 41.83 & 83.72 & 72.85 & 69.43\\
        LLaMA 3 & 44.97 & 17.86 & 39.54 & 83.64 & 64.56 & 62.29 & 45.27 & 19.36 & 41.47 & 81.52 & 68.35 & 66.72\\ \rowcolor{orange!30}
        \textbf{LLaMA 3 + HERA} & 48.53 & 21.26 & 42.73 & 88.26 & 74.18 & 73.46 & 50.45 & 23.75 & 44.16 & 88.27 & 79.32 & 78.59\\
        Gemini 1.5 & 45.28 & 18.39 & 40.25 & 84.35 & 65.17 & 62.56 & 45.71 & 19.62 & 41.33 & 81.27 & 67.82 & 65.94\\ \rowcolor{orange!30}
        \textbf{Gemini 1.5 + HERA} & \textbf{49.53} & \textbf{21.74} & \textbf{43.83} & \textbf{88.92} & \textbf{76.81} & \textbf{76.22} & \textbf{50.78} & \textbf{24.36} & \textbf{45.25} & 89.53 & \textbf{80.92} & \textbf{80.17}\\
        GPT-4 & 44.85 & 17.39 & 39.47 & 82.57 & 65.39 & 62.53 & 45.39 & 19.43 & 41.57 & 81.64 & 69.81 & 67.52\\ \rowcolor{orange!30}
        \textbf{GPT-4 + HERA} & 48.72 & 21.37 & 43.16 & 88.75 & 74.39 & 73.25 & 50.62 & 24.16 & 44.95 & \textbf{89.71} & 80.33 & 79.41\\
    \bottomrule
    \end{tabular}}
    \caption{Automatic evaluation results of HERA for factual consistency, relevance and fluency. \textbf{R-1/2/L} are ROUGE-1/2/L, \textbf{BS} is BERTScore, \textbf{FC} is FactCC, \textbf{SC} is SummaC. The best result per metric for each dataset is \textbf{bolded}.}
    \label{tab:results}
\end{table*}

\subsection{Experimental Settings}

\paragraph{Datasets} We evaluate the performance of HERA on two long document summarization datasets \textbf{arXiv} and \textbf{PubMed} \citep{cohan-etal-2018-discourse}. These two datasets contain academic papers in different scientific fields and are longer than commonly-used news datasets. We randomly select 500 articles from test set of each dataset to construct our test set, respectively.

\paragraph{Baselines} To validate the effectiveness of HERA, we apply HERA to four state-of-the-art LLMs, including \textbf{LLaMA 2 13B} \footnote{\href{https://llama.meta.com/llama2/}{https://llama.meta.com/llama2/}}, \textbf{LLaMA 3 8B} \footnote{\href{https://llama.meta.com/llama3/}{https://llama.meta.com/llama3/}}, \textbf{Gemini 1.5} \footnote{\href{https://ai.google.dev/gemini-api}{https://ai.google.dev/gemini-api}} and \textbf{GPT-4-0613} \footnote{\href{https://openai.com/index/gpt-4/}{https://openai.com/index/gpt-4/}}. 

We also compare HERA with recent long document summarization models: \textbf{FactorSum} \citep{fonseca-etal-2022-factorizing}, a factorized energy-based abstractive model that improves the performance and applicability by separate budget decisions from selecting important content in the document, and \textbf{Lodoss} \citep{cho-etal-2022-toward}, an extractive architecture that learns robust sentence representations by performing summarization and segmentation simultaneously. 

\paragraph{Metrics} We evaluate the factual consistency, fluency and informativeness of summaries using four different automatic metrics: (1) \textbf{FactCC} \citep{kryscinski-etal-2020-evaluating}, a weakly-supervised, model-based approach for verifying factual consistency and identifying conflicts between source documents and generated summaries, (2) \textbf{SummaC} \citep{laban-etal-2022-summac} that enables natural language inference models to detect inconsistency, (3) \textbf{ROUGE}\cite{lin-2004-rouge}, an automatic evaluation metric for the informativeness and fluency of a summary based on lexical overlap, and (4) \textbf{BERTScore} \citep{Zhang2020BERTScore}, that computes a similarity score between candidate and reference summaries using BERT contextual embeddings.

\paragraph{Implementation} We run open LLMs LLaMA 2 13B and LLaMA 3 8B with Text Generation Inference\footnote{\href{https://github.com/huggingface/text-generation-inference}{https://github.com/huggingface/text-generation-inference}} on 8 24GB NVIDIA GeForce RTX 3090 GPUs. HERA use BRIO \citep{liu-etal-2022-brio} to generate summary sentence of every paragraph and selects the Top 6 paragraphs to form a segment bag. Regarding other baselines used in the experiments, we use standard checkpoints provided by the authors and adopt the same configuration as in the corresponding papers, respectively.

\subsection{Main Results}

The experimental results of HERA and baselines on arXiv and PubMed are reported in Table \ref{tab:results}. For arXiv, Gemini 1.5 + HERA achieves a relative gain of 8.8\% on ROUGE-1 and 17.9\% on FactCC respectively. Importantly, Gemini 1.5 + HERA performs the best on all metrics compared with other baselines, showing that HERA significantly improves faithfulness and fluency of summaries. For PubMed, Gemini 1.5 + HERA and GPT-4 + HERA achieve almost the same excellent performance, and also outperform other baselines on both fluency and faithfulness metrics. 

Overall, LLMs combined with HERA uniformly outperforms foundation models themselves with a wide margin on all metrics, which demonstrates the generality and effectiveness of HERA. Specially, among four LLMs, HERA achieves the most outstanding performance on Gemini 1.5 and the greatest gain on LLaMA 2. Therefore, our approach can significantly improve the overall quality of summaries generated by various LLMs without requiring additional training and resources.

\subsection{Ablation Study}

In order to investigate the effect of context packaging and reordering, we conduct ablation studies using LLaMA 3. As shown in Table \ref{tab:ablation}, context packaging improves the performance of LLM on long document summarization. because it condenses useful information and removes potentially confusing information, preventing LLM from being misled and distracted by irrelevant content. Besides, context reordering can further enhance the quality of generated summaries, which proves that a good narrative order can significantly enhance LLM's understanding and utilization of context, and bring non-negligible performance improvements to LLM. Ablation experiments demonstrate the effectiveness and impact of context packaging and reordering on performance of HERA.

\begin{table}[t]
\centering
\resizebox{\linewidth}{!}{
    \begin{tabular}{l|cccc|cc}
    \toprule
         & \large{\textbf{R-1}} & \large{\textbf{R-2}} & \large{\textbf{R-L}} & \large{\textbf{BS}} & \large{\textbf{FC}} & \large{\textbf{SC}}\\ 
    \midrule
        \large{\textbf{Method}} & \multicolumn{6}{c}{\large{\textbf{arXiv}}}\\
    \midrule
        LLaMA 3 & 44.97 & 17.86 & 39.54 & 83.64 & 64.56 & 62.29\\ \rowcolor{orange!30}
        w packaging & 47.25 & 19.96 & 41.53 & 86.74 & 68.81 & 68.46\\
        w both & 48.53 & 21.26 & 42.73 & 88.26 & 74.18 & 73.46\\
    \midrule
        \large{\textbf{Method}} & \multicolumn{6}{c}{\large{\textbf{PubMed}}}\\
    \midrule
        LLaMA 3 & 45.27 & 19.36 & 41.47 & 81.52 & 68.35 & 66.72\\ \rowcolor{orange!30}
        w packaging & 48.72 & 22.85 & 43.29 & 86.64 & 74.52 & 73.61\\
        w both & 50.45 & 23.75 & 44.16 & 88.27 & 79.32 & 78.59\\
    \bottomrule
    \end{tabular}}
    \caption{The results of ablation experiments. We remove context reordering by directly generating summaries of every segment bag without sorting paragraphs.}
    \label{tab:ablation}
\end{table}

\subsection{Impacts of Hyperparameters}

We conduct quantitative experiments to investigate impacts of bag size using LLaMA 3. We varied the number of paragraphs selected for retrieval and evaluated changes in quality of summaries. Intuitively, shorter contexts can enable LLM to utilize the information in them more accurately, but retaining only too few paragraphs may lose key information for generating summaries. As can be seen, the trends of the results of the two datasets are not monotonous and similar. For example, table \ref{tab:bagsize} shows the fluency and faithfulness scores significantly increase when bag size increases from 3 to 5 for arXiv, but ROUGE-1, BERTScore and FactCC scores will decrease when bag size greater than 5.The quantitative experimental results show that LLM requires enough information to generate summaries, but a too large bag size will make the performance of HERA degenerate to the original model.

\begin{table}[t]
\centering
\resizebox{\linewidth}{!}{
    \begin{tabular}{l|ccc|ccc}
    \toprule
         & \large{\textbf{R-L}} & \large{\textbf{BS}} & \large{\textbf{FC}} & \large{\textbf{R-L}} & \large{\textbf{BS}} & \large{\textbf{FC}} \\ 
    \midrule
        \large{\textbf{Top-k}} & \multicolumn{3}{c|}{\large{\textbf{arXiv}}} & \multicolumn{3}{c}{\large{\textbf{PubMed}}}\\
    \midrule
        k = 3 & 32.94 & 73.52 & 51.76 & 35.83 & 73.64 & 56.19\\
        k = 4 & 37.41 & 81.35 & 63.59 & 40.68 & 82.45 & 70.53\\
        k = 5 & \cellcolor{orange!30} 42.73 & \cellcolor{orange!30} 88.26 & \cellcolor{orange!30} 74.18 & 44.16 & 88.27 & 79.32\\
        k = 6 & 42.58 & 88.13 & 73.92 & \cellcolor{orange!30} 44.57 & \cellcolor{orange!30} 88.63 & \cellcolor{orange!30} 79.38\\
        k = 7 & 41.29 & 86.82 & 71.68 & 43.61 & 86.72 & 75.65\\
        k = 8 & 40.68 & 84.57 & 67.49 & 41.59 & 82.94 & 70.48\\
    \bottomrule
    \end{tabular}}
    \caption{The performance of HERA with varying bag size k.}
    \label{tab:bagsize}
\end{table}

\begin{table}[htbp]
\centering
    \begin{tabular}{l|cc}
        \toprule
            \textbf{Method} & \textbf{arXiv} & \textbf{PubMed}\\
        \midrule
            LLaMA 2 & 60.57 & 37.46\\ \rowcolor{orange!30}
            LLaMA 2 + HERA & 74.43 & 54.05\\
            LLaMA 3 & 36.93 & 25.66\\ \rowcolor{orange!30}
            LLaMA 3 + HERA & 59.87 & 38.29\\
        \bottomrule
    \end{tabular}
    \caption{The inference time (minute) of two LLMs with and without HERA.}
    \label{tab:time}
\end{table}

\subsection{Time Cost}

We recorded the inference time of the LLaMA 2 13B and LLaMA 3 8B on two datasets to evaluate the computational cost of HERA. Table \ref{tab:time} shows two LLMs with HERA only took about 1.5 times as long as themselves without HERA. Although HERA adds two steps, because HERA generates local summaries in parts, it reduces the computational complexity and time cost compared to directly generating summaries for the entire long document. Therefore, HERA slightly increases the computational cost, but brings relatively high performance gains.

\section{Conclusion}

In this paper, we propose \textbf{HERA}, a novel LLM summary generation framework, which improves fluency, informativeness and faithfulness of long document summarization via context packaging and reordering without additional training and resources. We evaluate HERA on two popular benchmark datasets using four LLMs, and extensive experiments demonstrate that HERA can significantly improves the ROUGE, BERTScore, FactCC and SummaC scores of summaries generated by LLMs. Furthermore, we investigate the effect of context packaging, reordering and hyperparameters in HERA. We also evaluate the inference time of HERA to demonstrate that HERA only slightly increases the computational cost.

\section*{Limitations}

Although HERA improves the performance of foundation models on long document summarization, our approach does not optimize the prompt template for the subtask. Moreover, HERA does not use more powerful retrieval methods in context packaging. Besides, limited by the computational resources and
budget, we only evaluate our approach on a total of 1000 documents on two datasets and lack human evaluation.

Although our approach significantly improves the faithfulness of summaries, but the summaries generated by HERA may still contain misleading, distorted, and fake information, because the hallucination phenomenon of LLMs is difficult to eliminate.

\bibliography{custom}

\appendix

\section{Prompts}

Table \ref{tab:prompts} shows the prompts for subtasks of HERA.

\begin{table}[htbp]
\centering
    \resizebox{\linewidth}{!}{
    \begin{tabular}{p{\linewidth}}
    \toprule
        \rowcolor{gray!30} \textbf{Summary Generation}: Generate summary sentence for every paragraph.\\
    \midrule
        \underline{\textit{Instruction}}: Summarize the following paragraph in one sentences.\\
    \bottomrule
        \rowcolor{gray!30} \textbf{Paragraphs Retrieve}: Retrieve paragraphs related to an event.\\ 
    \midrule
        \underline{\textit{Instruction}}: Rank the following sentences based on their relevance to the event.\\
    \bottomrule
        \rowcolor{gray!30} \textbf{Event Extraction}: Extract important events from the original document.\\ 
    \midrule
        \underline{\textit{Instruction}}: Extract the most important events from the following summary sentences.\\
    \bottomrule
       \rowcolor{gray!30} \textbf{Summary Aggregation}: Aggregate all local summaries into an overall summary.\\ 
    \midrule
        \underline{\textit{Instruction}}: Generate connectives to concatenate all summaries to form a fluent text. DO NOT change the original semantics.\\
    \bottomrule
    \end{tabular}}
    \caption{Prompts used in HERA.}
    \label{tab:prompts}
\end{table}

\section{Experiments Details}

\subsection{Datasets}

\begin{table}[htbp]
    \centering
    \resizebox{\linewidth}{!}{
    \begin{tabular}{cccccccc}
    \toprule
        \multirow{2}{*}{\textbf{Dataset}} & \multirow{2}{*}{\textbf{Train}} & \multirow{2}{*}{\textbf{Valid}} & \multirow{2}{*}{\textbf{Test}} & \multicolumn{2}{c}{\textbf{Document}} & \multicolumn{2}{c}{\textbf{Summary}}\\
        & & & & \textbf{Words} & \textbf{Sents} & \textbf{Words} & \textbf{Sents}\\
    \midrule
         arXiv & 203,037 & 6,436 & 6,440 & 4,938 & 204.8 & 230.8 & 9.6\\
         PubMed & 119,224 & 6,633 & 6,658 & 3,049 & 86.43 & 203 & 6.8\\
    \bottomrule
    \end{tabular}}
    \caption{Statistics of summarization datasets used in this paper.}
    \label{tab:datasets}
\end{table}

Table \ref{tab:datasets} reports the statistics of two open-source datasets arXiv and PubMed that licensed under Apache 2.0, where Sents is Sentences. We randomly select 500 articles from standard test set of each dataset to construct our test set, respectively.

\subsection{Software and Licenses}

Table \ref{tab:software} lists the version and licenses of software used in this paper.

\begin{table}[htbp]
    \centering
    \resizebox{\linewidth}{!}{
    \begin{tabular}{lcc}
    \toprule
        \textbf{Software} & \textbf{Version} & \textbf{Licence}\\
    \midrule
         numpy & 1.21.5 & BSD \\
         torch & 1.13.1 & BSD-3\\
         NLTK & 3.7.0 & Apache 2.0\\
         sentencepiece & 0.1.96 & Apache 2.0\\ 
         text-generation-inference & 1.4.2 & Apache 2.0\\ 
         huggingface-hub & 0.11.1 & Apache 2.0\\
         datasets & 2.12.0 & Apache 2.0\\
         transformers & 4.36.1 & Apache 2.0\\
         rouge-score & 0.1.2 & Apache 2.0\\
         bert-score & 0.3.13 & MIT\\
         factcc & / & BSD 3\\
         summmac & 0.0.4 & Apache 2.0\\
         LLaMA & 2 & Llama 2 Community\\
         LLaMA & 3 & Llama 3 Community\\
         Gemini & 1.5 & Proprietary\\
         GPT & 4 & Proprietary\\
    \bottomrule
    \end{tabular}}
    \caption{Version and licenses of software used in this paper.}
    \label{tab:software}
\end{table}

\end{document}